\documentclass[letterpaper, 10 pt, conference]{ieeeconf}  
\IEEEoverridecommandlockouts
\usepackage{cite}
\usepackage{amsmath,amssymb,amsfonts}
\usepackage{algorithmic}
\usepackage{graphicx}
\usepackage[hyphens,spaces,obeyspaces]{url}
\usepackage{hyperref}
\usepackage{todonotes}
\usepackage{textcomp}
\usepackage{xcolor}
\usepackage{bm}
\usepackage{threeparttable}
\usepackage{siunitx}
\usepackage{multirow}
\usepackage{comment}
\usepackage{booktabs}
\usepackage{comment}

\def\BibTeX{{\rm B\kern-.05em{\sc i\kern-.025em b}\kern-.08em
    T\kern-.1667em\lower.7ex\hbox{E}\kern-.125emX}}

\hypersetup{
	colorlinks=true,
	linkcolor=blue,
	filecolor=magenta,      
	urlcolor=blue,
}

\begin{document}


\title{\LARGE \bf{The Effect of Anthropomorphism on Trust in an Industrial Human-Robot Interaction}
}

\author{Tim~Schreiter$^1$$^*$,
       Lucas Morillo-Mendez$^2$$^*$,
       Ravi T. Chadalavada$^1$,
       Andrey Rudenko$^3$, \\
       Erik Alexander Billing$^4$,
       and~Achim J.~Lilienthal$^1$
\thanks{$^{1}$Mobile Robotics and Olfaction Lab,
	\"Orebro University, Sweden {\tt\small \{tim.schreiter, ravi.chadalavada, achim.lilienthal\}@oru.se}}
\thanks{$^{2}$Machine Perception and Interaction Lab,
	\"Orebro University, Sweden {\tt\small lucas.morillo@oru.se}}
\thanks{$^{3}$Robert Bosch GmbH, Corporate Research, Stuttgart, Germany
{\tt\small andrey.rudenko@de.bosch.com}}%
\thanks{$^{4}$Interaction Lab, University of Skövde, Sweden 
{\tt\small erik.billing@his.se}}%
\thanks{$^{*}$These authors contributed equally to the work.}
\thanks{This work was supported by the European Union’s Horizon 2020 research and innovation program under grant agreement No. 101017274 (DARKO) and grant agreement No. 754285}}

\maketitle

\begin{abstract}
    Robots are increasingly deployed in spaces shared with humans, including home settings and industrial environments. In these environments, the interaction between humans and robots (HRI) is crucial for safety, legibility, and efficiency. A key factor in HRI is trust, which modulates the acceptance of the system. Anthropomorphism has been shown to modulate trust development in a robot, but robots in industrial environments are usually not anthropomorphic. We designed a simple interaction in an industrial environment in which an anthropomorphic mock driver (ARMoD) robot simulates driving an autonomous guided vehicle (AGV). The task consisted of a human crossing paths with the AGV, with or without the ARMoD mounted on the top, in a narrow corridor. The human and the system needed to negotiate trajectories when crossing paths, meaning that the human had to attend to the trajectory of the robot to avoid a collision with it. There was a significant increment in the reported trust scores in the condition where the ARMoD was present, showing that the presence of an anthropomorphic robot is enough to modulate the trust, even in limited interactions as the one we present here.
\end{abstract}

\section{Introduction}\label{sec:related}

\begin{figure}[!t]
\centering
\includegraphics[width= 0.8\linewidth]{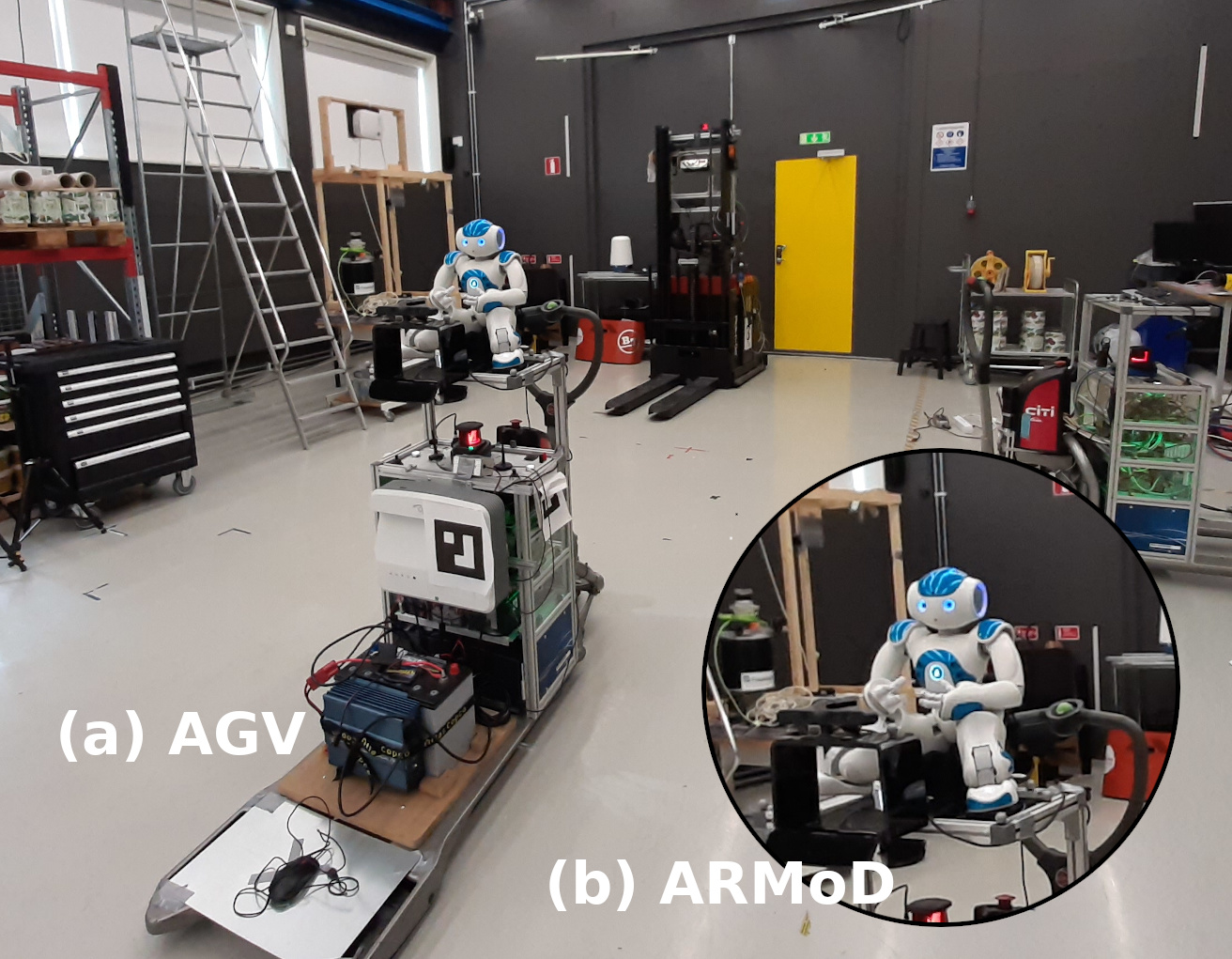}
\caption{Setup of the (a) Autonomous Guided Vehicle (AGV) with the (b) Anthropomorphic Robotic Mock Driver (ARMoD), placed on top of the vehicle. The AGV can potentially communicate intent through gestures, gazes and speech of the Driver.}
\label{abb:naosetup}
\end{figure}

Non-humanoid robots are usually equipped with various non-verbal channels to communicate their intent to humans, such as light signals, floor projections, or auditory signals \cite{cha2018survey}. When navigating in a shared environment, such signals help coordinating their motion with people, to avoid collision, increase legibility and task efficiency. As for the tasks that require cooperation and active coordination, like the handover task, more complex communication channels might be required, such as verbal communication with additional gestures and gazes. Humanoid robots with anthropomorphic features, e.g., arms, legs, and facial features, are often used in this context to improve interaction with human users \cite{szabov20}, \cite{homburg21}, and their anthropomorphism leads to an increase of users' trust in the interaction \cite{natarajan2020effects, Christoforakos2021}.

As the usage and complexity of industrial robots increase, they take on unfamiliar shapes and thus, complicate the interaction and establishment of trust in shared environments with humans. Robot-related factors were shown to be the most relevant for the development of trust in these interactions \cite{hancock2011meta}. Among these factors, the design of the robot is particularly important to get the human interaction partners to trust the robot appropriately \cite{kok2020trust}. Anthropomorphic features may aid the trust, but they are not often present in industrial robots. As the complexity of the new systems increase, the perception of these systems as collaborators rather than machines have been deemed as positive \cite{Stadler2013}. For example, the addition of a pair of sunglasses to a industrial robotic hand and gripper, along with a set of breathing-like movements and gaze behavior, improved metrics from participants such as the perceived sociability and likeability of the system \cite{Terzi20}. However, in their study, the authors did not find differences in trust. One possible reason for this is that they included a scale that was not originally designed for industrial collaborations \cite{Heerink2009}. Another study showed that trust does not seem to be affected as a result of anthropomorphism in industrial settings \cite{Onnasch2021}. In this case the authors used a validated scale to trust in industrial collaboration, developed by Charalambous et al.\cite{charalambous2016development}. Nevertheless, the study employed a limited form of anthropomorphism in which a face appeared on a screen attached to a robotic arm and gripper. In contrast to previous research, our study does not include a tactile interaction. Our research explores a new perceived modality of navigation for a non-humanoid robot that can potentially improve trust on the system by using a humanoid robot, NAO, in industrial settings. 

 We propose a combined approach of a humanoid robot with an Autonomous Guided Vehicle (AGV), used for instance in the intralogistic settings\footnote{\url{http://iliad-project.eu}}. We refer to this combination as a ``robot-on-robot platform'' (see Figure \ref{abb:naosetup}). By combining an AGV with sophisticated social robots with anthropomorphic features, successfully used in trust-related user studies \cite{ueno2022trust}, we expect to achieve an increase in the trust of human users. To the best of our knowledge, this is the first approach to study the interaction of an navigating AGV equipped with an an Anthropomorphic Mock Driver (ARMoD) with participants in a shared environment. To this extent, we chose an NAO robot with a human-likeliness score of 46\% \cite{phillips18}, as it is the subject of recent user based studies \cite{amirova21}, it is small enough to be mounted on our AGV, and posses a software development kits to develop custom modules. We designed an encounter in a narrow corridor to measure the impact of the ARMoD on trust reported by participants. In our scenario, as the participant approached the robot, it looked at the participant's head and traced it until the vehicle passed the participant. We used the scale developed by Charalambous et al.\cite{charalambous2016development} to measure trust in industrial collaborations. With the data we obtained from 33 participants, we found that the users reported higher trust in the interaction with the ARMoD than in the AGV alone.

\section{Experiment design}\label{subsec:encounter}
\label{sec:experiments_description}\label{sec:methods}

\subsection{Materials}

The aim of this study was to explore the impact that an Anthropomorphic Robotic Mock Driver (ARMoD) seated on top of an AGV has on users' trust during a basic human-robot interaction consisting of walking and avoiding the moving platform. The AGV we used fits the definition of ``mobile platform'' after ISO8373:2021. On top of the platform a seat was mounted to hold the NAO robot in place, which enabled a fixed and repeatable placement (see Figure  \ref{abb:naosetup}). We used the same AGV as in our previous studies of intent communication \cite{chadalavada2020bi}, a retrofitted Linde CitiTruck with a SICK S300 scanner at the back, to ensure the safety of humans approaching from behind. The AGV is equipped with sensor modules to localize itself within the laboratory and a onboard RGB-D camera for short-range person detection ($\approx $2 meters).

\subsection{Task}
We designed a scenario where participants encountered a moving robot platform in a hallway with the aim to study trust as a result of the appearance of the platform, either as it is (AGV), or with a anthropomorphic robot on the top (ARMoD). In the experiment we chose a setup that reflects a potential encounter between humans and robotic workers in an industrial environment. The chosen width of the hallway was 2 meters, as it matches the regulations for corridors proposed in the DIN-18040-1 and the EN-ISO-24341 (former EN 426) standards for meeting areas. The participants and the platform started 14 meters apart, a feasible length for the definition as a corridor encounter. During this sequence, the participants saw the platform as it approached them, and they were simply instructed to walk by its side towards the opposite direction. In the ARMoD condition, if the participants got close enough for the short-range person detection ($\approx $2 meters), the robot simulated awareness with the head motion to trace the participants' movement until they crossed paths. This encounter was repeated three times with the platform randomly taking one of the different routes: (1) platform moved in a curve to the right side of the hallway, (2) platform moved to the left side of the hallway and (3) platform headed straight. The platform moved at a constant speed of 0.6 m/s.

\begin{figure}[t]
\centering
\includegraphics[width= 1.1\linewidth]{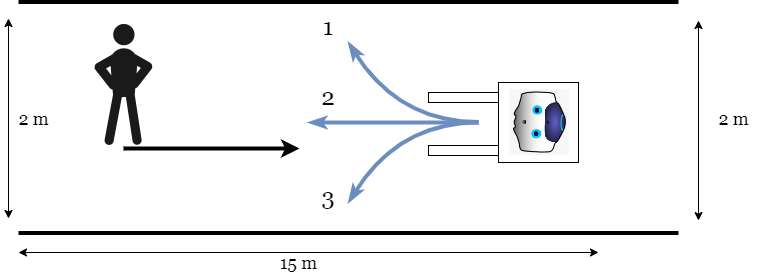}
\caption{Participant encountering the robot-on-robot platform in a 2-meter wide and 15-meter long corridor. The participant has to decide for a side to pass the platform. The robot takes one of three different trajectories \textbf{(1)} Curve to the right side, \textbf{(2)} Straight ahead, \textbf{(3)} Curve to the left side.}
\label{abb:expsetupcorr}
\end{figure}

\subsection{Procedure, sample, and measurements}
After the task, participants filled an adapted version of a scale to measure trust in industrial human-robot collaboration developed by Charalambous et al. \cite{charalambous2016development}. In this version, the items referring to the grip of the robot (C, E, G, J) were removed, as no gripper was used in our scenario. We also used a 7-point Likert scale ranging from 1 ``strongly disagree'' to 7 ``strongly agree''.  Additionally, we obtained some demographic information (see Table \ref{tab:demo}). The age of participants ranged from 18 to 56 years (M=28.7, SD=7.88) and all of them were fluent in English. Participants were recruited at Örebro University and participation was voluntary. All participants were informed about the task, consented to participate, and were aware of the possibility to leave at any time. We analyzed the trust scores of 33 participants divided in two groups: one that walked by the ARMoD on top of the AGV (n=19) and one that walked just by the AGV (n=14). 

\begin{table}[t]
\centering
\caption{Demographic information from the participants}
\label{tab:demo}
\begin{tabular}{llllll}
\toprule
Group & N & Age (SD) & Women & Other gender & Left-handed \\
\midrule
ARMoD & 19 & 29.7 (9.8) & 13 & 1 & 1 \\ 
AGV & 14 & 27.3 (4) & 5 & 0 & 1 \\ 
\bottomrule
\end{tabular}
\end{table}

\section{Results}\label{sec:discuss}

To ensure that the trust scale of Charalambous et al. \cite{charalambous2016development} was still reliable despite the removal of certain items, we calculated the Cronbach's $\alpha$. The scale yielded a score of 0.76, beyond the acceptable level of 0.7 \cite{Nunnally1994}. \\ 
The trust scale is composed of three major components, the robot's motion and pickup speed (1), safe co-operation (2), and robot and gripper reliability (3). Because the first and last components involve the gripper and the pickup action, which were not part of our experiment, we just used those items within these that applied to the robot but not to the gripper: one item for the first component (robot's motion and pickup speed, two items in the original scale), and one item for the third component (robot and gripper reliability, four items in the original scale). To calculate the final trust score, we multiplied each of these item's score by the number of items belonging to that component in the original scale, two and four respectively, and added these to the sum of the scores of the second component (safe co-operation). 

\begin{figure}[t]
      \centering
      \includegraphics[width = 1\linewidth]{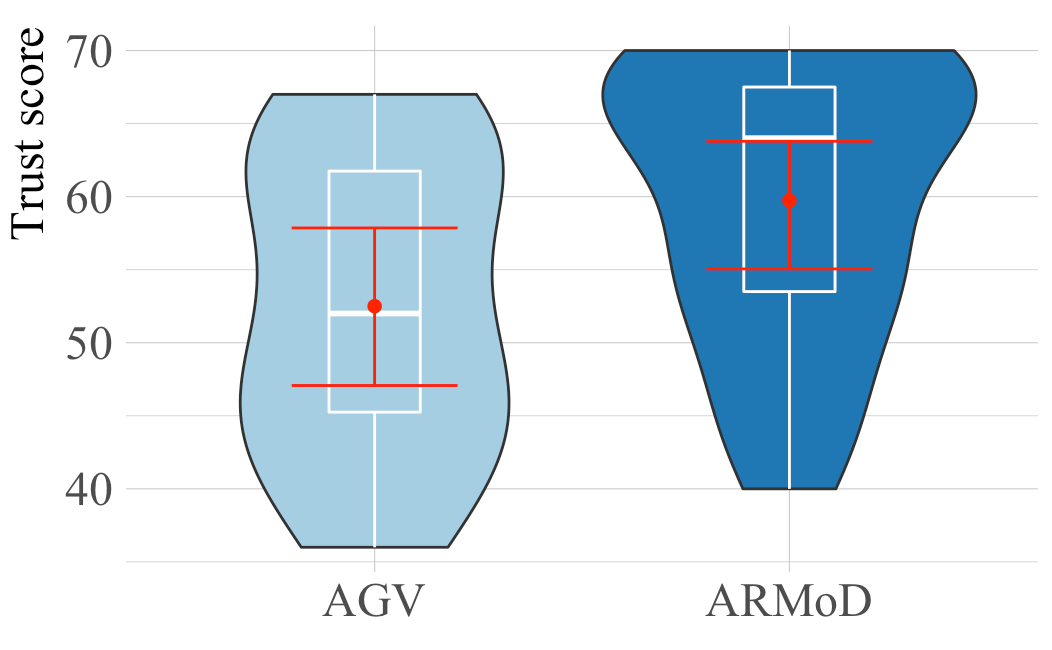}
       \caption {Violin plots and box plots of the trust scores for both conditions. Means and corresponding error bars in red. Error bars show 95\% bootstrapped confidence intervals.}
      \label{trust}
\end{figure}

Once the trust score was obtained for each participant, we proceeded with the analysis, performed in R \cite{Rsoft}. Because the trust scores in both groups were not normally distributed (see violin plots in Fig. \ref{trust}), we opted for a Robust variation of the Welch's t-test \cite{Welch1947} to compare the reported trust between the two groups. The \textit{yuenbt} function, based on bootstrapping, from the WRS2 \cite{Wilcox2016, Mair2020} package was used for the analysis. We kept the default bootstrapping value of 599 samples of 20\% trimmed means. The $\hat{\xi}$ measure was used as a explanatory measurement of robust effect size, as suggested by Wilcox and Tian \cite{Rand2011}. This measure was calculated using the \textit{yuen.effect.ci} function of the WRS2 package. Values of $\hat{\xi}$= 0.1, 0.3, and 0.5 correspond to small, medium and large effect sizes respectively.

On average, participants reported higher levels of trust for the ARMoD condition (M = 59.73, SE = 2.22), than for the AGV alone (M = 52.5, SE = 2.78). This difference was
marginally significant $t=-1.68, p=.051, 95\% CI[-17, 0.09];$ nevertheless, this difference did represent a medium–large effect, $\hat{\xi}=0.41$.

\section{Discussion}

In this study we explored how trust from users varied as a result of the anthropomorphic features of a robot in an industrial environment. Hancock et al. \cite{hancock2011meta} outlined the importance of robot-related factors for the development of trust in a human-robot interaction. In \cite{kok2020trust}, the authors state that improving trust in a robot already starts with designing it appropriately. Instead of designing a new robot from scratch to increase the trust of users in industrial settings, we modified the design of an ``Autonomous Guided Vehicle'' (AGV) by adding an ``Anthropomorphic Robotic Mock Driver'' (ARMoD). AGVs such as the forklift in our experiment are frequently deployed in shared industrial environments alongside human co-workers. We used the popular anthropomorphic social robot ``NAO'' \cite{amirova21, ueno2022trust} as the ARMoD. 

The results of our study showed that the use of the ARMoD increased the reported trust in the interaction of participants with the platform. The simple addition of a robot on the top of the AGV, alongside basic gaze behavior, was enough to increase users' trust of the system within an industrial setting. Our results are in line with recent research emphasizing the role of anthropomorphism in trust \cite{natarajan2020effects, Christoforakos2021}. Contrary to other research set in industrial environments \cite{Terzi20, Onnasch2021}, our results showed that perceived trust varies as a result of anthropomorphism in basic interactions such as the avoidance of a moving robot. Although not complex, this kind of interaction is probably one that will be common in busy industrial settings. The difference in the results between this study and prior art is probably explained by the different nature of the interactions, as previous studies involved tasks such as handovers and precise object manipulations in which the success of the interaction may have not been taken for granted by participants. 

Our suggested solution increased trust through the use of anthropomorphic features and gaze behavior. However, there are other features that can lead to set an appropriate level of trust during the interactions with humans. For example, we previously explored a different method of communicating intent for the AGV using ``Spatial Augmented Reality (SAR)'' by projecting patterns on the floor in front of the robot \cite{chadalavada2020bi}. This form of communication is however limited by the lighting conditions of the environment and can only be deployed to communicate navigational intent. Using the ARMoD we can overcome the disadvantages of the previous studied SAR to design future experiments, e.g., independence from the lighting conditions or two dimensional floor patterns. The ARMoD can interact with participants in a proactive manner through the usage of non-verbal communication, as well as gazes and gestures to communicate any kind of intent. Future research should explore how these social features beyond plain anthropomorphism might impact the trust that users have on the robots in industrial environments. 

This research comes with two limitations. First, although high levels of trust are desirable, we just focused on the robot appearance component that modulates it. Appropriate functioning and the minimization of failures by the system can have a greater impact on the perceived trust. Moreover, manipulating trust purely by appearance while ignoring other aspects could lead to over-trust, which can be dangerous and is not desirable in potentially threatening situations, such as the platform not breaking when headed towards a person. Second, we designed a basic encounter that did not involve a tactile interaction or manipulation, contrary to previous research with industrial robots. Nevertheless, we believe that the scenario of a corridor encounter with a robot will likely become an everyday common one. This is because this situation may occur in a wide variety of industries, and other kind of shared environments and with different types of workers, even with those not directly involved in close collaborative works with the robot. 

\section{Conclusion and Future Work}\label{sec:concl}\label{sec:Future}

This work presented a study on a novel interaction method for an autonomous guided vehicle (AGV) through an anthropomorphic robotic mock driver (ARMoD) with 33 participants. We conducted an experiment, a hallway encounter, where participants passed the AGV with the ARMoD mounted on top in a narrow corridor in one condition and encountered only the AGV, but without the ARMoD in another. We found that through adding the ARMoD to the AGV, participants reported a higher trust on the interaction. Our results suggest that using anthropomorphic robot in industrial settings can help to adjust the levels of trust that users place on what otherwise would be a navigation vehicle. Under the assumption of proper functioning, higher trust on industrial robots can lead to higher levels of acceptability reported by the users. 

\section{Acknowledgement}

We want to acknowledge Timm Linder and Chittaranjan Swaminathan for their help in setting up the software, Per Sporrong for his help with setting up the hardware and Per Lindström for manufacturing the seat of the mock driver used in this study.

\bibliographystyle{IEEEtran}
\bibliography{IEEEabrv,references}

\end{document}